\documentclass[10pt,twocolumn,letterpaper]{article}

\usepackage{iccv}
\usepackage{times}
\usepackage{epsfig}
\usepackage{graphicx}
\usepackage{amsmath}
\usepackage{amssymb}
\usepackage{caption}
\usepackage{booktabs}
\usepackage{multirow}
\usepackage{siunitx}
\usepackage{subcaption,amsfonts,dcolumn}
\usepackage{dblfloatfix}
\newcolumntype{d}[1]{D..{#1}}

\usepackage[breaklinks=true,bookmarks=false]{hyperref}

\iccvfinalcopy 

\def\iccvPaperID{****} 

\ificcvfinal\fi

\begin{document}

\title{MOTS R-CNN: Cosine-margin-triplet loss for multi-object tracking}

\author{Amit Satish Unde and  Renu M. Rameshan\\
School of Computing and Electrical Engineering, Indian Institute of Technology, Mandi, India\\
{\tt\small amitsunde@gmail.com, renumr@iitmandi.ac.in}
}

\maketitle
\ificcvfinal\thispagestyle{empty}\fi

\begin{abstract}
One of the central tasks of multi-object tracking involves learning a distance metric that is consistent with the semantic similarities of objects. Design of an appropriate loss function that encourages discriminative feature learning is among the most crucial challenges in deep neural network based metric learning. Despite significant progress, slow convergence and a poor local optimum of the existing contrastive and triplet loss based deep metric learning methods necessitates a better solution. In this paper, we propose cosine-margin-contrastive (CMC) and cosine-margin-triplet (CMT) loss by reformulating both contrastive and triplet loss functions from the perspective of cosine distance. The proposed reformulation as a cosine loss is achieved by feature normalization which distributes the learned features on a hypersphere. We then propose the MOTS R-CNN framework for joint multi-object tracking and segmentation, particularly targeted at improving the tracking performance. Specifically, the tracking problem is addressed through deep metric learning based on the proposed loss functions. We propose a scale-invariant tracking by using a multi-layer feature aggregation scheme to make the model robust against object scale variations and occlusions. The MOTS R-CNN\footnote{Our code will be made available at: \textcolor{red}{https://github.com/amitsunde/MOTS-R-CNN}}  achieves the state-of-the-art tracking performance  on the KITTI MOTS dataset. We show that the MOTS R-CNN reduces the identity switching by $62\%$ and $61\%$ on cars and pedestrians, respectively in comparison to Track R-CNN. 

\end{abstract}

\section{Introduction}
Multi-object tracking (MOT) is a well-established problem in the field of computer vision \cite{mot1, mot2, mot3, mot4}. It is regaining attention of the research community owing to significant progresses in autonomous vehicles and robotics \cite{trajectory1, trajectory2, trajectory3}. The “tracking-by-detection” is the widely adopted strategy to tackle MOT problem \cite{mot3, amir, trackrcnn}. It is the process of detecting objects of interest in each video frame and matching their identities across frames to obtain the object trajectories over time. Given a video frame, the object detector identifies and locates possible objects of interest and feeds them as input to the tracking system. The tracker computes the similarity score between the feature vectors of previously tracked targets and objects detected in the current frame. Based on the similarity score, the tracker links detections and targets to find optimal object trajectories.
\par 
The traditional tracking system consists of two parts: 1) robust hand-crafted feature extraction for each detected object \cite{old1, old2, old3} and 2) distance metric learning consistent with the semantic similarities of objects \cite{distance, deepmetric1, deepmetric3}. After an exhaustive study, it is noticed that the discrimination power of hand-crafted features is low that results in sub-optimal performance \cite{deepmetric2, oldargue, oldargue1}. With the evolution of convolutional neural networks (CNNs) in recent years, deep CNN-based object tracking has gained increasing popularity \cite{mot4, amir, attention}. Different from traditional methods, CNN-based approaches learn the feature representation of each detected object and distance metric in an end-to-end manner. MOT problem using CNNs has been thoroughly studied over the past few years which in turn results in a noticeable improvement in the tracking performance \cite{realtimemot, uma, graphcnn}. However, tracking remains a challenging problem, especially in unconstrained scenarios such as autonomous driving, due to severe occlusions, scale and appearance variation of objects, false positives (FP) and missing detections,  and illumination variations \cite{amir, oldargue1, motsnet}. 

One possible reason that affects the tracking performance of MOT is the use of axis-aligned rectangular bounding boxes \cite{siamese, bbox}.  This choice of the bounding box may contain information about the background and other objects in close proximity due to the overlapping of bounding boxes, especially in crowded scenes, which deteriorates the efficiency of the tracker. Hence, approaches to improve the tracker performance is shifting from the bounding box level to the pixel level.  In this direction, Voigtlaender et al. proposed a unified framework for multi-object tracking and segmentation (MOTS) through the integration of the tracking head in the Mask R-CNN model \cite{maskrcnn}. Most of the approaches thereafter reported in the literature are focused on learning embedding representation of objects using triplet loss \cite{trackrcnn, motsnet, remots}. However, existing deep metric learning frameworks aiming to learn discriminative features based on triplet loss often suffer from slow convergence and a poor local optimum \cite{deepmetric1,deepmetric2,deepmetric3,deepmetric4}. 

\par 
In this paper, we propose a cosine-margin-triplet (CMT) and cosine-margin-contrastive (CMC) loss to enhance the feature discrimination power of the multi-object tracking model. Since the Euclidean distance between the feature vectors can be unbounded, we utilize the dot product between the pair of normalized feature vectors which is equal to the cosine similarity. The feature normalization distributes the learned feature vectors on a hypersphere and encourages better feature learning by separating them in angular. In addition, a margin term is introduced in the cosine metric to increase the intra-class compactness and inter-class discrimination. 

\par 
We then propose the MOTS R-CNN network to address the MOTS problem. The proposed model extends the Mask R-CNN \cite{maskrcnn} with the addition of the tracking head. The 3D convolutional layers are integrated into the MOTS R-CNN framework to model the spatio-temporal dependency of objects. The tracking problem is addressed through joint learning of feature representation and distance metric based on the CMT loss in an end-to-end manner.  The tracking capability of the MOTS R-CNN to handle object scale variations is improved by fusing features from different layers of the convolutional network. Further improvement in tracking accuracy is achieved by taking advantage of the readily available instance segmentation mask in the MOTS R-CNN to focus only on foreground information of objects. In addition, we empirically show the inefficiency of the softmax loss for object classification. We demonstrate that incorporating large margin cosine loss (LMCL) \cite{cosface} in the MOTS R-CNN increases the object classification accuracy. 
The major contributions of our work are as follows:
\begin{itemize}
	\item We propose cosine-margin-triplet and cosine-margin-contrastive loss functions to enforce closeness between pairs of feature vectors belonging to the same instance while pushing dissimilar vectors far away on the embedding space. The proposed loss functions encourage discriminative feature learning for multi-object tracking problems.  
	\item We present the MOTS R-CNN model to address the MOTS problem. We show that identity switching (IDS) using the proposed model for cars and pedestrians is reduced by $62\%$ and $61\%$ respectively in comparison with Track R-CNN that uses triplet loss for distance metric learning.
	\item 	We empirically demonstrate the usefulness of large margin cosine loss over traditional softmax loss for object classification.
\end{itemize}

The remainder of this paper is organized as follows. We discuss in section \ref{related work} the existing works related to MOT and MOTS problem. In section \ref{distance metric learning}, we review preliminaries for distance metric learning. In section \ref{CMT}, we propose the CMT and CMC loss functions. We present the proposed MOTS R-CNN model in section \ref{MOTSRCNN}. The effectiveness of the proposed model through experimental analysis is shown in section \ref{experiments} and followed by conclusions in section \ref{conclusions}. 

\section{Related Work}\label{related work}
There are only few works reported in the literature to address the MOTS task since it is introduced very recently. So, we first discuss existing literature related to the MOT problem. Then, we briefly review the related work on the MOTS task.

\noindent \textbf{Multi-object tracking.}
The work in \cite{amir} combined appearance, motion, and interaction cues using a recurrent neural network to make the network robust against occlusions. However, a large number of IDS as compared to state-of-the-art methods show that the combination of various cues does not guarantee an improvement in the tracking performance. Ji Zhu et al. proposed a spatial and temporal attention model to extract only object-specific features and to reduce the effect of noisy observations on the tracking performance \cite{attention}.  Their work was mainly focused on the extension of a single object tracker for the MOT task. As an alternative to the traditional tracking-by-detection approach, a joint detection and tracking method was proposed in \cite{realtimemot} by integrating the tracking head in the single-shot detector to develop the real-time MOT system. While the joint paradigm speed up the overall process, the inferior performance of the single-shot detector caused a reduction in the tracking accuracy. 

\par 
The work in \cite{uma} extended Siamese single object tracker for the MOT task through the integration of motion estimation and data association into a single framework.  The combination of appearance and motion features from 2D and 3D space jointly was proposed in \cite{graphcnn} for improving the accuracy of the tracker. Furthermore, a graph neural network was employed to promote feature interaction of various objects aiming at more discriminative feature learning.  Sarthak Sharma et al. proposed the BeyondPixels model for MOT by taking the 3D pose, shape, and motion information of objects into account \cite{beyondpixels}. The matching was done only in the specific search area obtained through the propagation of position and orientation information of objects in successive frames.  

\par 
\noindent \textbf{Multi-object tracking and segmentation.} Paul Voigtlaender et al. proposed the Track R-CNN framework to integrate object detection, instance segmentation, and tracking in a unified framework \cite{trackrcnn}. The tracking head was incorporated in parallel to the box and mask heads of the Mask R-CNN architecture. The tracking problem was posed as a feature learning problem by minimizing the triplet loss. It was observed that the tracking accuracy was poor both in the case of cars and pedestrians. This can be due to the use of triplet loss for distance metric learning and the extraction of features only from the final convolutional layer of the backbone network. This single layer feature extraction can make the tracking performance sensitive to the object scale variation. 

\par 
The MOTSNet proposed in \cite{motsnet} achieved the scale-invariant feature learning by making use of the feature pyramid network \cite{fpn} to form the backbone convolutional network. The segmentation mask was used to improve the tracking performance by extracting only the foreground information of objects. The triplet loss was employed to facilitate discriminative feature learning for object tracking. However, the overall performance improvement was largely attributed to training the network with their automatically generated MOTS training dataset. The tracking-by-points strategy was proposed in \cite{pointrack} by interpreting image pixels as irregular 2D point clouds. The shape, color, semantic class labels, and position information of objects was taken into account for the object tracking.  The incorporation of motion information coupled with the recovery of missing detections in the MOTS framework was proposed in \cite{motsfusion}. While MOTSFusion yields promising tracking performance, the two-stage tracking process involving the construction of short tracks using optical flow and its projection into a 3D space makes it computationally heavy. The computational overhead in MOTSFusion can restrict its extension to real-time applications such as autonomous driving.

\section{Review of distance metric learning}\label{distance metric learning}
In this section, we review various loss functions that are used for distance metric learning.

\noindent \textbf{Softmax loss.} The softmax loss is widely used for object classification, face recognition, and person re-identification \cite{sphereface, arcface, softmaxreid1, softmaxreid2}. It enables feature learning by formulating the learning task as a multi-class classification problem. The softmax loss is given as,

\begin{equation}
L=\frac{1}{N}\sum_{i=1}^{N}-log\frac{e^{W_{y_i}^T x_i} + b_{y_i}}{\sum_{j=1}^{n}e^{W_{j}^T x_i} + b_{j}}
\end{equation}
where $(x_i, y_i)$ denotes the input feature vector to the softmax layer and associated class label respectively and $N$ is the number of training examples. The terms $W_j$ and $b_j$ represent the $j^{th}$ column of the weight matrix $W$ and corresponding bias term respectively. In spite of its wide use, the learned features using the softmax loss function are known to be less discriminative \cite{cosface, arcface, sphereface}.

\noindent \textbf{Large margin cosine loss.} The LMCL overcomes limitations of the traditional softmax loss \cite{cosface}. It forces the network to distribute learned features on a hypersphere through $\ell_2$ normalization of $W_j$ and $x_i$. The LMCL introduces a positive margin $m$ in the loss function and is expressed as,
\begin{equation}
L=\frac{1}{N}\sum_{i=1}^{N}-log\frac{e^{s(\cos(\theta_{y_i, i})-m)}}{e^{s(\cos(\theta_{y_i, i})-m)}+\sum_{j \neq y_i}e^{s\cos(\theta_{j, i})}}
\end{equation} 
where $\theta_{j, i}$  is the angle between normalized $W_j$ and $x_i$. The incorporation of LMCL in face recognition models lead to noticeable performance improvement \cite{cosface}.

\noindent \textbf{Contrastive loss.} Given a pair of feature vectors $(f_{i}, f_{j})$ together with the binary label $(y_{ij})$, the contrastive loss minimizes the distance between the feature vectors belonging to the same class while penalizing the distance between negative pair if it is smaller than a specified margin $m$ \cite{deepmetric1, deepmetric2}. The contrastive loss is defined as,
\begin{equation}
L=y_{ij} \parallel f_{i}-f_{j}\parallel_2^2 + (1-y_{ij}) \max (0, m - \parallel f_{i}-f_{j}\parallel_2)^{2}
\end{equation} 
where $y_{ij}=1$ when $(f_{i}, f_{j})$ are from the same class and $y_{ij}=0$ when $(f_{i}, f_{j})$ belongs to the different class.

\noindent \textbf{Triplet loss.} Triplet loss has gained a lot of attention for person re-identification and object tracking tasks due to its superior performance in deep face recognition \cite{facenet, triplet, triplet1, trackrcnn, motsnet}. It is defined on triplets $(f_{i}^a, f_{i}^p, f_{i}^n)$, where $f_{i}^a$ is referred to as anchor of the triplet such that positive pair $(f_{i}^a, f_{i}^p)$ have the same class label and the negative pair $(f_{i}^a, f_{i}^n)$ have the different label. Triplet loss distributes the learned feature vectors on an embedding space where the distance between positive pair is larger than that of negative pair by a specified margin $m$. It is given as,
\begin{equation}
L = \max(0, \parallel f_{i}^a - f_{i}^p \parallel_2^2 - \parallel f_{i}^a - f_{i}^n \parallel_2^2 + m )
\end{equation} 
\par The performance of both contrastive and triplet loss based distance metric learning methods depends on the hard sampling which is employed to obtain nontrivial pairs or triplets in large mini batches. Despite their widespread use in various applications, both the loss functions have drawbacks of slow convergence and poor local optima \cite{deepmetric1,deepmetric2,deepmetric3,deepmetric4}.

\section{Proposed CMT and CMC loss functions}\label{CMT}
One of the important tasks of multi-object tracking involves learning a distance metric that is consistent with the semantic similarities of objects. Distance metric learning is the process of learning an embedded representation of the objects to keep the distance between similar objects at the minimum while pushing away instances of dissimilar objects far on the embedding space. In this section, we propose a cosine-margin-triplet loss and cosine-margin-contrastive loss to enhance the feature discrimination power of the multi-object tracking model. 

\subsection{Cosine-margin-triplet loss}
Motivated from the success of triplet loss for face recognition and person re-identification, we design the loss function based on triplets of feature vectors obtained using deep CNNs.  Let $D$ denote the set of detections over a batch of video frames. For each detected object $d_i \in D$, there is an associated ground truth segmentation mask, class label, tracking identity, and an association feature vector $f_i$.  We define a loss function on the triplets $(f_i, f_i^+, f_i^-)$, where $f_i$  is the anchor of the triplet that shares the same tracking identity with the anchor positive $(f_i^+)$  and has a different identity (but the same class) with the anchor negative $(f_i^-)$. The loss function is formulated as,
\begin{equation}\label{triplet1}
L = \frac{1}{N}\sum_{i=1}^{N}-log\frac{e^{f_i^T f^+}}{e^{f_i^T f^+} + e^{f_i^T f^-}}
\end{equation}
where $N=|D|$. The loss function in Eq. \eqref{triplet1} aims to minimize the positive pair, $(f_i, f_i^+)$ distance, and penalize the negative pair, $(f_i, f_i^-)$ distance. It builds upon the dot product between the pair of feature vectors which is given as,
\begin{equation}
f_i^T f_i^+=\parallel f_i \parallel \parallel f_i^+ \parallel \cos(\theta_i^+)
\end{equation}
where $\theta_i^+$ is the angle between $f_i$ and  $f_i^+$.

The numerical value of the dot product is influenced by both the direction and norm of feature vectors. In order to develop effective feature learning, it is essential to make the dot product determined only by the direction. Hence, the feature vectors are normalized to have unit norm by $\ell_2$  normalization. In order to enable sufficient angular separability, the radius of the sphere is scaled to $s$. This step distributes the learned feature vectors on a hypersphere of radius $s$. The cosine similarity between the normalized feature vectors has a geometric correspondence to the geodesic distance on the hypersphere.  This correspondence can be attributed to the relationship between the central angle and arc of a circle and also to the fact that the learned embeddings are angular separable. The normalized loss function is expressed as, 
\begin{align}
L &= \frac{1}{N}\sum_{i=1}^{N}-log\frac{e^{s(f_i^T f^+)}}{e^{s(f_i^T f^+)} + e^{s(f_i^T f^-)}} \\
&= \frac{1}{N}\sum_{i=1}^{N}-log\frac{e^{s(\cos(\theta_i^+))}}{e^{s(\cos(\theta_i^+))} + e^{s(\cos(\theta_i^-))}}\label{triplet2}
\end{align}
where $f_i = \frac{f_i}{\parallel f_i \parallel}$, $f_i^+ = \frac{f_i^+}{\parallel f_i^+ \parallel}$, and $f_i^- = \frac{f_i^-}{\parallel f_i^- \parallel}$. 

However, the features learned using the above loss function are not guaranteed to be highly discriminative. It does not take the stringent constraints for discrimination into account which can produce ambiguity in decision boundaries. For example, let $\theta_i^+$ and $\theta_i^-$ denote the angle between the positive pair (same tracking identity) and negative pair (different tracking identity) respectively. The normalized loss function in Eq. \eqref{triplet2} forces $\cos(\theta_i^+) > \cos(\theta_i^-)$ to map similar feature vectors close to each other.
\par 
To further enhance the intra-class compactness and inter-class discrimination, we introduce the margin term in the cosine metric. To be more specific, the loss function is modified to force $\cos(\theta_i^+)-m > \cos(\theta_i^-)$, where the hyperparameter $m$ is the additive cosine margin. Since  $\cos(\theta_i^+)-m$  is less than $\cos(\theta_i^+)$, the incorporation of the cosine margin penalty in the loss function makes the constraints more stringent and thereby promotes discriminative feature learning.

The cosine-margin-triplet loss that enforces the distribution of positive pair closer on the hypersphere while pushing away negative pair is defined as, 
\begin{equation}
L_{CMT} = \frac{1}{N}\sum_{i=1}^{N}-log\frac{e^{s(\cos(\theta_i^+) - m)}}{e^{s(\cos(\theta_i^+) - m)} + e^{s(\cos(\theta_i^-))}}
\end{equation}
\par The tracking model is trained by sampling hard positive and hard negative for each detected object.

\begin{figure*}[t]
	\centering
	\includegraphics[width=0.9\linewidth]{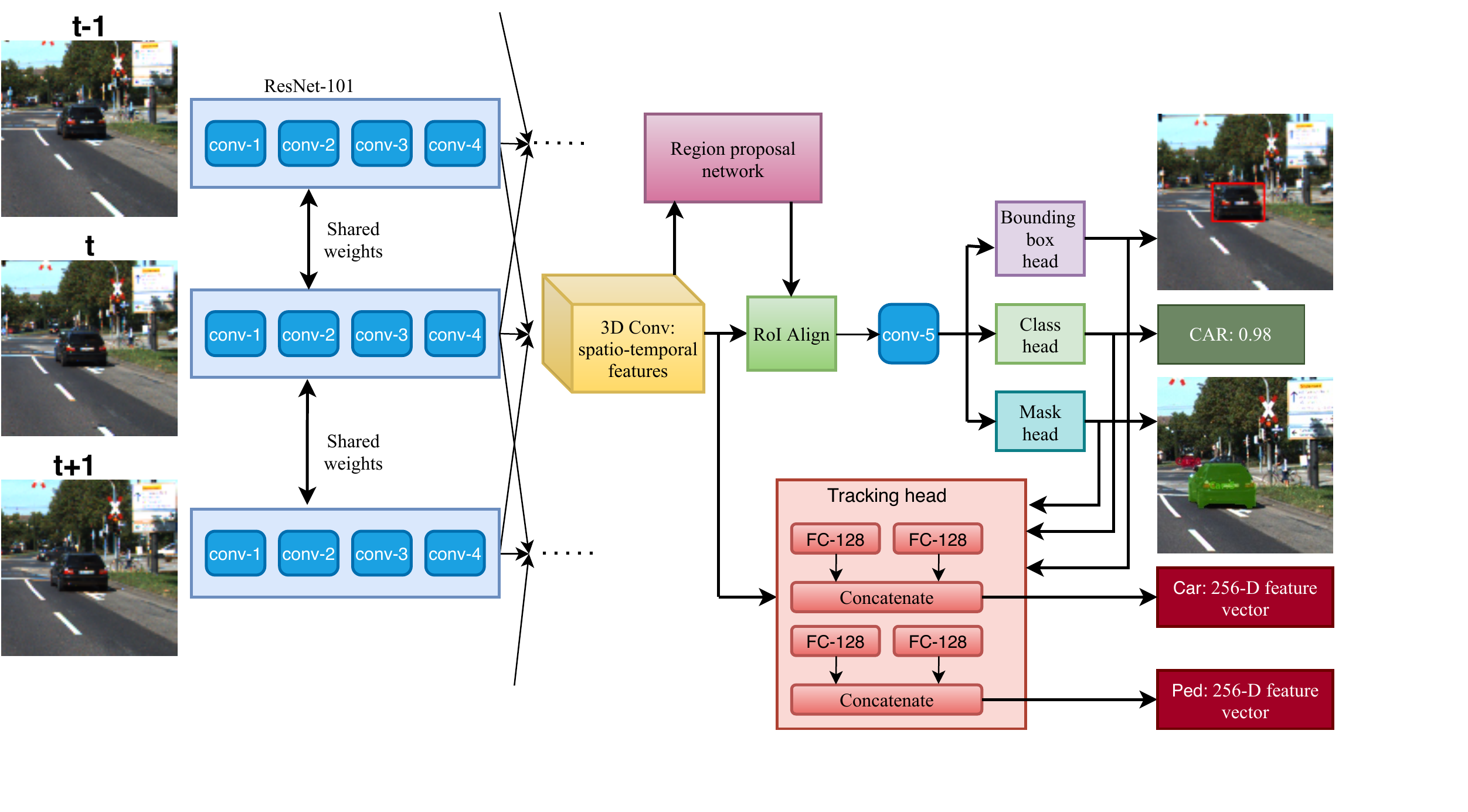}
	\caption{Overview of the proposed MOTS R-CNN. The nomenclature conv-x denotes the depth of the backbone network.}
	\label{fig:architecture}
\end{figure*}  

\subsection{Cosine-margin-contrastive loss}
While the CMT loss results in the relative distance, we propose the loss function that measures the absolute distance, identical to the contrastive loss, to answer the question “How similar/dissimilar are the feature vectors of two objects?” The proposed cosine-margin-contrastive loss is defined as, 

\begin{multline}
L_{CMC} = \frac{1}{N}\sum_{i=1}^{N}(-loge^{\sigma (s(\cos(\theta_i^+)-m))} \\
-loge^{1-\sigma(s(\cos(\theta_i^-)-m))}) \label{final}
\end{multline} 
where $f_i = \frac{f_i}{\parallel f_i \parallel}$, $f_i^+ = \frac{f_i^+}{\parallel f_i^+ \parallel}$,  $f_i^- = \frac{f_i^-}{\parallel f_i^- \parallel}$, $N=|D|$, and $\sigma(.)$ is the sigmoid activation function. It is worth noting that the application of the sigmoid function bounds the cosine similarity score to $[0, 1]$ and hence $1- \sigma(.)$ gives cosine distance between pair of feature vectors.  The first term $-loge^{\sigma (s(\cos(\theta_i^+)-m))}$ in Eq. \eqref{final} is responsible for mapping the positive pair closer while the second term $-loge^{1-\sigma(s(\cos(\theta_i^-)-m))})$ is responsible for pushing the negative pair far away on the hypersphere.

The loss function in Eq. \eqref{final} can be easily reformulated as two-class softmax loss and is given as, 
\begin{multline}
L_{CMC} = \frac{1}{N}\sum_{i=1}^{N}(-log\frac{e^{\sigma (s(\cos(\theta_i^+)-m))}}{e^{\sigma (s(\cos(\theta_i^+)-m))}+e^{1 - \sigma (s(\cos(\theta_i^+)))}} \\
-log\frac{e^{1-\sigma(s(\cos(\theta_i^-)-m))}}{e^{1-\sigma(s(\cos(\theta_i^-)-m))} + e^{\sigma(s(\cos(\theta_i^-)))}}) \label{final_soft}
\end{multline}  

Similar to the CMT loss, hard positive and hard negative sampling is used during training.

\section{Proposed MOTS R-CNN}\label{MOTSRCNN}
We address the MOTS problem building on the strengths of Mask R-CNN that extends the Faster R-CNN detector \cite{fasterrcnn} with an instance segmentation (mask) head. We propose MOTS R-CNN which is an extension of Mask R-CNN with the addition of a tracking head. The architecture of MOTS R-CNN has 1) a bounding box and classification head for object localization and classification, 2) a mask head for instance segmentation of objects, and 3) a tracking head for data association.  Given an image, MOTS R-CNN outputs a set of detections with associated bounding boxes, class labels, and instance segmentation masks together with learned feature vectors.  These learned embeddings are then used to associate detected objects in the video sequence.

\subsection{Overview}
The proposed MOTS R-CNN architecture is shown in Fig. \ref{fig:architecture}. The first 91 convolutional layers (up to conv-4) in ResNet-101 \cite{resnet} is used as the convolutional backbone network.  The 3D convolutional layers are applied to the feature maps obtained from the backbone network. The incorporation of 3D convolutions efficiently models the spatio-temporal dependency of objects \cite{3dconv}.  These spatio-temporal features are then shared by a region proposal network (generating a set of object proposals) and a Fast R-CNN detection network \cite{fastrcnn}. RoI-align is performed on spatio-temporal features to extract a portion of the feature map specified by object proposals. All convolutional layers of conv-5 in ResNet-101 are applied on RoI-aligned features which are shared by box, classification, and mask head. Similar to the Mask R-CNN, the box head, classification head, and mask head localizes the objects of interest, assigns a class label, and generates binary segmentation masks respectively. The detected objects augmented with their class labels, and segmentation masks are feed as input to the tracking head for the association. 

\subsection{Tracking head}
MOTS R-CNN extends the original implementation of Mask R-CNN through the incorporation of the tracking head in series with the box, classification, and mask heads. This cascade arrangement of the tracking head is encouraged to speed up the network during inference due to the processing of the fewer number of boxes and to improve accuracy by using more accurate bounding boxes. The tracking head consists of fully-connected layers with bounding boxes, class labels, and masks as inputs and an association feature vector for each detected box as output.  It is worth noting that occlusion, noisy detections, and appearance changes pose several inevitable challenges in multi-object tracking scenarios. Besides, the scale of various objects also affects the tracking performance adversely. For example, the size of an occluded pedestrian is small relative to a non-occluded car. To address the aforementioned challenges, multi-layer feature aggregation is used in the proposed work as described in the following. 

\noindent \textbf{Multi-layer feature aggregation for scale-invariant tracking.}
We first extract the object-specific feature maps from spatio-temporal features using RoI-align for each detected object. On these RoI-aligned features, all layers of conv-5 are applied.  We refer to the feature map corresponding to the final convolutional layer of 3D convolutions and conv-5 in ResNet-101 as C4 and C5 respectively. The RoI-aligned features are likely to be comprised of both foreground and background information due to the axis-aligned rectangular bounding boxes. It is well known from recent studies that foreground information alone contributes significantly to object tracking. Hence, it is highly essential to ensure that the tracking performance is largely dependent on foreground information. 

\par 
Since the MOTS framework is equipped with instance segmentation masks, foreground-background discrimination is readily available.  In view of the fact that C4 has high spatial resolution than C5, we pixel-wise multiply RoI-aligned features from C4 with the corresponding mask to nullify the effect of background information. Finally, the RoI-align masked features and C5 features are independently converted to 128-dimensional feature vectors by fully-connected layers. These individual 128-dimensional feature vectors are then concatenated to obtain a 256-dimensional embedding vector for each detected object.   These feature vectors are then learned using cosine-margin-triplet loss. 

\noindent \textbf{Data Association.}
At the test time, the dot product between normalized feature vectors of detected objects gives the similarity score which is used for data association. Specifically, we link each detected object in the current frame $t$ having a detection confidence score greater than a threshold $\beta$ with detections in the previous frames. The detections from previous $t-\alpha$ frames are considered for tracking. The detections in the current frame are compared with previously detected objects if and only if (1) corresponding bounding box centers have Euclidean distance less than $\gamma$ and (2) associated pairwise similarity score is larger than the threshold $\delta$.  Matching is performed using a greedy algorithm and all high confidence detections that are not linked to any previously detected objects are assigned to a new track.

\subsection{Training MOTS R-CNN}
The MOTS R-CNN is trained jointly in an end-to-end manner by defining multi-task loss on each detected object. This is done through the addition of the losses associated with the box, classification, mask, and tracking head together. The multi-task loss is given as,
\begin{equation}
L = L_{BH} + L_{LMC} + L_{MH} + L_{CMT}
\end{equation}
where $ L_{BH}$ and $L_{MH}$ denote the losses corresponding to the box head and mask head respectively as defined in \cite{maskrcnn}. Different from the Mask R-CNN, we use LMCL and CMT loss for classification and tracking head respectively.    

\section{Experiments}\label{experiments}
In this section, we report a detailed experimental analysis to test the effectiveness of the MOTS-RCNN and thereby the proposed loss functions on the KITTI MOTS dataset \cite{trackrcnn}. We carry out ablation studies to signify the contribution of each component of the MOTS R-CNN. We present the performance of the proposed algorithm in comparison with the state-of-the-art methods for the MOTS task. The performance is evaluated in terms of sMOTSA, MOTSA, IDS, FP, and false negatives (FNs) metrics as described in \cite{trackrcnn}. While the sMOTSA and MOTSA measure the overall accuracy of the MOTS network, IDS evaluates the performance of the tracking head. The FP and FN measure the performance of the bounding box and classification head.

\begin{table*}[t]
	\caption{Performance of the MOTS R-CNN on KITTI validation dataset in comparison with Track R-CNN}  
	\centering
	\begin{tabular}{|l|l|l|l|l|l|l|l|l|l|l|}
		\hline
		\multirow{2}{*}{Method} &
		\multicolumn{2}{c|}{sMOTSA $\uparrow$} &
		\multicolumn{2}{c|}{MOTSA $\uparrow$} &
		\multicolumn{2}{c|}{IDS $\downarrow$} &
		\multicolumn{2}{c|}{FP $\downarrow$} &
		\multicolumn{2}{c|}{FN $\downarrow$} \\
		\cline{2-11}
		& {Car} & {Ped} & {Car} & {Ped} & {Car} & {Ped} & {Car} & {Ped} & {Car} & {Ped}\\
		\hline
			MOTS R-CNN & $\mathbf{78.2}$ & $\mathbf{49}$ & $\mathbf{90.1}$ & $\mathbf{67.3}$ & $\mathbf{35}$ & $\mathbf{30}$ & $\mathbf{130}$ & $\mathbf{208}$ & $\mathbf{629}$ & 856\\
		Track R-CNN \cite{trackrcnn} & 76.2 & 46.8 & 87.8 & 65.1 & 93 & 78 & 134 & 267 & 753 & $\mathbf{822}$ \\
		\hline
	\end{tabular}
	\label{tab:0}
\end{table*}

\subsection{Experimental Setup}
For the MOTS R-CNN, the ResNet-101 is used as the backbone convolutional network and it is pre-trained on ImageNet \cite{imagenet}, COCO \cite{coco}, and Mapillary \cite{map} dataset. Two depthwise separable 3D convolutional layers with filters of size $3 \times 3 \times 3$ followed by ReLU activation function are applied on the top of the backbone network.  The data augmentations including random flipping and gamma correction in the $[-0.2, 0.2]$ range are used during training. We train the MOTS R-CNN using the Adam optimizer \cite{adam} for $5$ epochs. The learning rate of the optimizer is set to $5 \times 10^{-5}$. The network is trained with mini-batches of size $4$ which are formed by stacking adjacent frames of the same video. 

\par 
During training, the hyper-parameters of the proposed cosine-margin-triplet loss including margin and scale $(m-s)$ are set to $0.15-8$ and $0-1$ for cars and pedestrians respectively. Afterwards, the tracking head for pedestrians is fine-tuned with mini-batches of size $4$ by setting $m$ and $s$ to $0.2-16$ and $0.25-8$ respectively. We then fine-tune tracking head for cars with mini-batches of size $8$ and $16$ respectively. We set the hyper-parameters for data association during the inference as follows: $\alpha$ equal to $10$ and $6$ for car and pedestrian, detection confidence score $\beta$ is set to $0.32$ and $0.345$ for car and pedestrian, $\gamma$ and $\delta$ are set to $150$ and $0.3$ respectively. All experiments are performed on the NVIDIA Quadro GP100 card with 16 GB of memory.

\subsection{Main Results}
We evaluate the effectiveness of the proposed MOTS R-CNN on the KITTI MOTS validation dataset. We compare the performance of the MOTS R-CNN with the existing Track R-CNN \cite{trackrcnn}, MOTSFusion \cite{motsfusion}, BeyondPixels \cite{beyondpixels}, and CIWT \cite{ciwt} models. The last two approaches are primarily designed for MOT task and extended to MOTS problem through the addition of the mask head as described in \cite{trackrcnn}. The performance of the tracking-by-detection paradigm heavily relies on detection results. Hence, we consider reported results with the same detection and segmentation pipeline as MOTS R-CNN to maintain fairness in the analysis.  

\noindent \textbf{Comparison with Track R-CNN.} The comparison of the proposed algorithm is given in Table \ref{tab:0}. To be more specific, our work differs from Track R-CNN in the three aspects 1) multi-layer feature aggregation mechanism, 2) the proposed CMT loss for the tracking head, and 3) the LMCL for object classification. While MOTS R-CNN gives $2\%$ sMOTSA improvement in the car class, the performance of pedestrians rises by $2.2\%$. The most encouraging finding is that ID switching using the proposed algorithm for both cars and pedestrians is reduced by $60\%$ in comparison with Track R-CNN which uses triplet loss for distance metric learning. This reduction in ID switching using MOTS R-CNN can be attributed to the proposed CMT loss and multi-layer feature aggregation mechanism. Specifically, it signifies that the proposed CMT loss function guides the model to converge to a better optimal minimum than the triplet loss.
Furthermore, $2.2\%$ improvement in MOTSA for both cars and pedestrians is due to the better classification accuracy. This demonstrates the prominence of large margin cosine loss in reducing the number of false positives. 

\noindent \textbf{State-of-the-art comparison.} We compare in Table \ref{tab:1} the performance of the MOTS R-CNN with the existing methods. While the MOTS R-CNN outperforms the BeyondPixels and CIWT, its performance is comparable with MOTSFusion \cite{motsfusion}. Different from MOTS R-CNN which uses appearance features, MOTSFusion exploits motion information for object tracking. It can be seen that the MOTS R-CNN keeps the number of IDS at the minimum, thereby attaining the state-of-the-art tracking performance. Furthermore, MOTSFusion has a large number of parameters since it requires an additional deep network for the computation of optical flow. It is worth noting that the reduced number of FNs for cars using the MOTS R-CNN $(629 \; \text{vs.}\; 673)$ indicates that the proposed algorithm tracks more number of ground truth objects, which is highly desirable. This validates the usefulness of the MOTS R-CNN model in practical applications. On the contrary, the improvement in sMOTSA and MOTSA for pedestrians using MOTSFusion is attributed to its ability to recover missing detections.  

\begin{table*}
	\caption{Performance of the MOTS R-CNN on KITTI validation dataset in comparison with existing algorithms}  
	\centering
	\begin{tabular}{|l|l|l|l|l|l|l|l|l|l|l|}
		\hline
		\multirow{2}{*}{Method} &
		\multicolumn{2}{c|}{sMOTSA $\uparrow$} &
		\multicolumn{2}{c|}{MOTSA $\uparrow$} &
		\multicolumn{2}{c|}{IDS $\downarrow$} &
		\multicolumn{2}{c|}{FP $\downarrow$} &
		\multicolumn{2}{c|}{FN $\downarrow$} \\
		\cline{2-11}
		& {Car} & {Ped} & {Car} & {Ped} & {Car} & {Ped} & {Car} & {Ped} & {Car} & {Ped}\\
		\hline
		MOTS R-CNN & $\mathbf{78.2}$ & 49 & $\mathbf{90.1}$ & 67.3 & $\mathbf{35}$ & $\mathbf{30}$ & 130 & 208 & 629 & 856\\
		MOTSFusion \cite{motsfusion}  & $\mathbf{78.2}$ & $\mathbf{50.1}$ & 90.0 & $\mathbf{68.0}$ & 36 & 34 & $\mathbf{94}$ & $\mathbf{181}$ & 673 & 855  \\
		BeyondPixels \cite{beyondpixels} & 76.9 & - & 89.7 & - & 88 & - & 280 & - & $\mathbf{458}$ & -  \\
		CIWT \cite{ciwt} & 68.1 & 42.9 & 79.4 & 61.0 & 106 & 42 & 333 & 401 & 1214 & 863 \\
		\hline
	\end{tabular}
	\label{tab:1}
\end{table*}

\begin{table*}[!]
	\caption{Ablation results on the KITTI MOTS validation dataset}
    \centering
	\begin{subtable}[t]{0.48\textwidth}
		\begin{tabular}{|l|l|l|l|l|l|l|}
			\hline
			\multirow{2}{*}{Method} &
			\multicolumn{2}{c|}{sMOTSA $\uparrow$} &
			\multicolumn{2}{c|}{MOTSA $\uparrow$} &
			\multicolumn{2}{c|}{IDS $\downarrow$} \\
			\cline{2-7}
			& {Car} & {Ped} & {Car} & {Ped} & {Car} & {Ped} \\
			\hline
			Multi-layer & 78.2 & 49 & 90.1 & 67.3 & $\mathbf{35}$ & $\mathbf{30}$ \\
			Single-layer & 77.8 & 47.7 & 89.8 & 66.6 & 39 & 50  \\
			\hline
		\end{tabular}
		\caption{Feature aggregation}
		\label{tab:ab_1}
	\end{subtable}
	\hspace{\fill}
	\begin{subtable}[t]{0.48\textwidth}
		
		\begin{tabular}{|l|l|l|l|l|l|l|}
			\hline
			\multirow{2}{*}{Method} &
			\multicolumn{2}{c|}{sMOTSA $\uparrow$} &
			\multicolumn{2}{c|}{MOTSA $\uparrow$} &
			\multicolumn{2}{c|}{IDS $\downarrow$} \\
			\cline{2-7}
			& {Car} & {Ped} & {Car} & {Ped} & {Car} & {Ped} \\
			\hline
			Greedy & 78.2 & 49 & 90.1 & 67.3 & $\mathbf{35}$ & $\mathbf{30}$ \\
			Hungarian & 78 & 48.6 & 90 & 66.8 & 47 & 46  \\
			\hline
		\end{tabular}
		\caption{Data association mechanisms}
		\label{tab:ab_2}
	\end{subtable}

	\begin{subtable}[t]{0.48\textwidth}
		\begin{tabular}{|l|l|l|l|l|l|l|}
			\hline
			\multirow{2}{*}{Loss function} &
			\multicolumn{2}{c|}{sMOTSA $\uparrow$} &
			\multicolumn{2}{c|}{FP $\downarrow$} &
			\multicolumn{2}{c|}{FN $\downarrow$} \\
			\cline{2-7}
			& {Car} & {Ped} & {Car} & {Ped} & {Car} & {Ped} \\
			\hline
			LMCL & 78.2 & 49 & 130 & $\mathbf{208}$ & $\mathbf{629}$ & 856 \\
			Softmax  & 76.9 & 48.2 & 125 & 232 & 835 & 850  \\
			\hline
		\end{tabular}
		\caption{Object classification loss functions}
		\label{tab:ab_3}
	\end{subtable}
	\hspace{\fill}
	\begin{subtable}[t]{0.49\textwidth}
		
		\begin{tabular}{|l|l|l|l|l|l|l|}
			\hline
			\multirow{2}{*}{Method} &
			\multicolumn{2}{c|}{sMOTSA $\uparrow$} &
			\multicolumn{2}{c|}{MOTSA $\uparrow$} &
			\multicolumn{2}{c|}{IDS $\downarrow$} \\
			\cline{2-7}
			& {Car} & {Ped} & {Car} & {Ped} & {Car} & {Ped} \\
			\hline
			With mask  & 78.2 & 49 & 90.1 & 67.3 & $\mathbf{35}$ & $\mathbf{30}$ \\
			Without mask  & 77.7 & 48.5 & 89.8 & 66.8 & 44 & 38  \\
			\hline
		\end{tabular}
		\caption{Instance segmentation mask for feature extraction}
		\label{tab:ab_4}
	\end{subtable}

	\label{tab:table1}
\end{table*}

\begin{table}
	\caption{Ablation results for distance metric learning}  
	\centering
	\begin{tabular}{|l|l|l|l|l|l|l|}
		\hline
		\multirow{2}{*}{Loss function} &
		\multicolumn{2}{c|}{sMOTSA $\uparrow$} &
		\multicolumn{2}{c|}{MOTSA $\uparrow$} &
		\multicolumn{2}{c|}{IDS $\downarrow$} \\
		\cline{2-7}
		& {Car} & {Ped} & {Car} & {Ped} & {Car} & {Ped} \\
		\hline
		CMT  & $\mathbf{78.2}$ & $\mathbf{49}$ & $\mathbf{90.1}$ & $\mathbf{67.3}$ & $\mathbf{35}$ & $\mathbf{30}$ \\
		CMC  & 77.8 & 48.5 & 89.8 & 67.2 & 46 & 34  \\
		\hline
	\end{tabular}
	\label{tab:dml}
\end{table}

\subsection{Ablation Studies} 
We perform several ablations to signify the importance of each module of MOTS R-CNN. 

\noindent \textbf{Multi-layer feature aggregation.}
We report in Table \ref{tab:ab_1} the effectiveness of the multi-layer feature aggregation over the single-layer feature extraction. More specifically, the features from the final convolutional layer of the ResNet-101 are converted to $128$-dimensional vectors by the fully connected layer and used for data association. The multi-layer feature aggregation significantly reduces the number of IDS. We observed empirically that the use of multi-layer features makes the network robust against occlusions.

\noindent \textbf{Data association mechanism.}
The performance of MOTS R-CNN using the popular Hungarian and greedy algorithms for data association is detailed in Table \ref{tab:ab_2}. The reduction in the number of IDS using the greedy algorithm indicates that the learned features using the proposed method are strongly discriminative. Also, the greedy algorithm is extremely lightweight in comparison to the Hungarian algorithm which speeds up the tracker at test time.   

\noindent \textbf{Object classification loss function.}
In Table \ref{tab:ab_3}, we demonstrate the significance of LMCL over the traditional softmax loss for the task of object classification. While the incorporation of LMCL in MOTS R-CNN reduces the number of FNs  from $835$ to $629$ for cars, it reduces the number of FPs from $232$ to $208$ in the case of pedestrians.

\noindent \textbf{Instance segmentation mask for feature extraction.} Table \ref{tab:ab_4} vividly illustrates the usefulness of the segmentation mask for extracting features from the foreground of objects to improve the tracking performance.  

\noindent \textbf{Distance metric learning.} 
We compare in Table \ref{tab:dml} the effect of the proposed CMT and CMC loss functions on feature learning for multi-object tracking. From an empirical analysis, the margin $m$ and scale factor $s$ are set to $0.35$ and $8$ for the CMC loss. It can be noticed that the performance using the CMT loss is better than the CMC loss. Specifically, the reduction in the number of IDS using the CMT loss illustrates its power to encourage discriminating feature learning. This observation is consistent with the fact that the traditional triplet loss which results in relative distance between pair of feature vectors is seen as an improvement over the contrastive loss.    

\begin{figure}[h]
	\centering
	\includegraphics[width=1\linewidth]{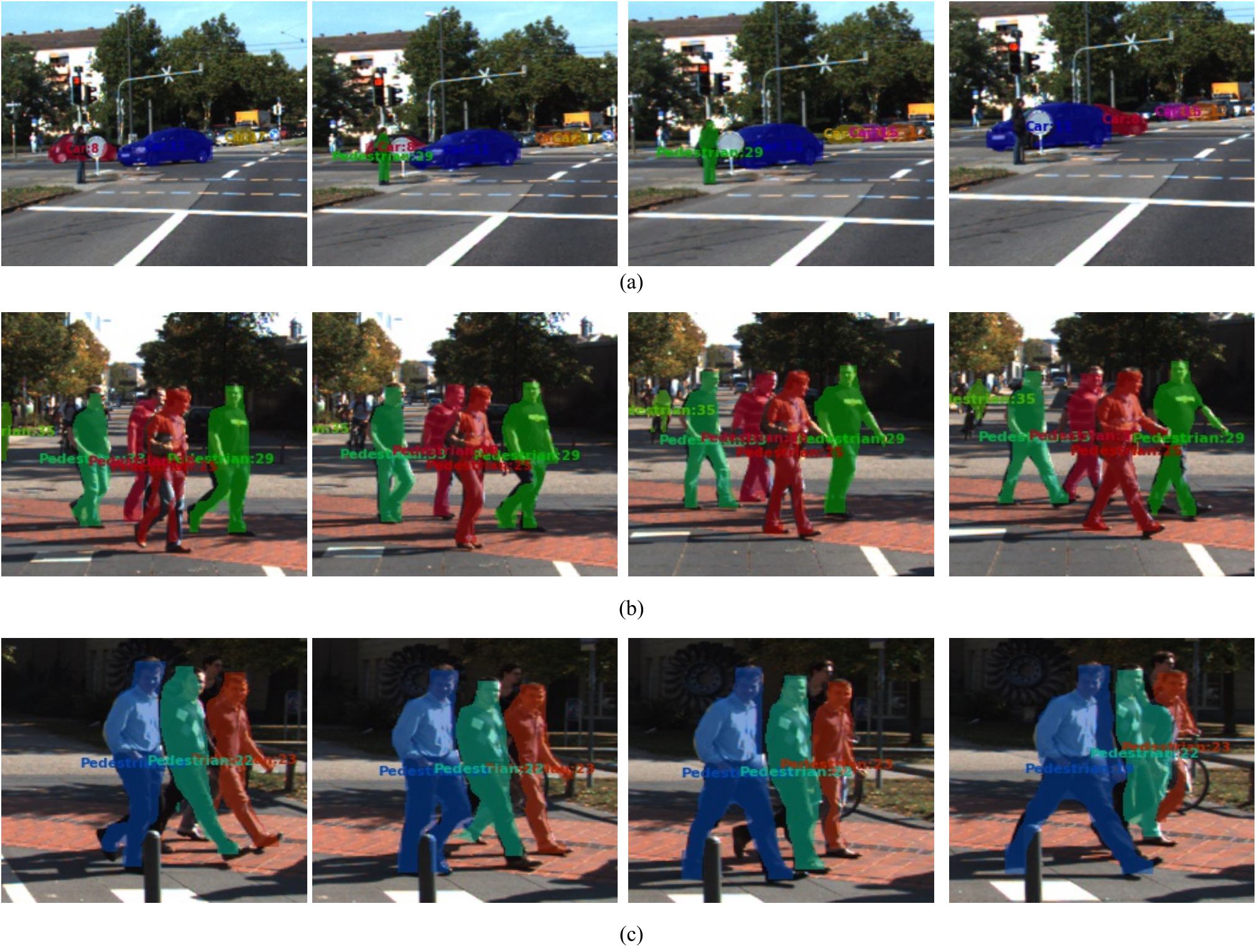}
	\caption{Qualitative analysis of tracking results on the KITTI MOTS validation dataset (a) robustness against merge and split problem and (b)+(c) robustness to occlusions}
	\label{fig:results}
\end{figure}

\noindent \textbf{Qualitative analysis.}
We display in Fig. \ref{fig:results} the qualitative tracking results for visual analysis. It can be noticed that our model is robust against merge and split problem. Furthermore, the ability of the model to handle occlusions and to perform consistent tracking is also witnessed. 

\section{Conclusions}\label{conclusions}
In this paper, we proposed cosine-margin-triplet and cosine-margin-contrastive loss functions for deep metric learning. While feature normalization makes features obtained via CNNs angular separable, the incorporation of margin term further boosts its discriminative ability. We then propose the MOTS R-CNN model to demonstrate the significance of the proposed loss functions for multi-object tracking. We show the robustness of our model to occlusions and scale variation which can be attributed to the multi-layer feature aggregation mechanism. The significant reduction in the number of IDS validates the better convergence of the proposed CMT loss. We believe that further improvement in the performance is possible by using a backbone convolutional network involving a feature pyramid network. 


\section*{}
\vspace*{-2em}
\noindent \textbf{Acknowledgement:} This work is funded by  SERB NPDF grant (ref. no. PDF/2019/003459), government of India.

{\small
\bibliographystyle{ieee_fullname}
\bibliography{egbib_1}
}

\end{document}